\documentclass[11pt]{article}

\usepackage[preprint]{acl}

\usepackage{times}
\usepackage{latexsym}
\usepackage{makecell}
\usepackage[T1]{fontenc}

\usepackage[utf8]{inputenc}
\usepackage{amsmath}
\usepackage{microtype}

\usepackage{inconsolata}

\usepackage{graphicx}
\usepackage{hyperref}
\usepackage{tcolorbox}

\title{Large Language Models Approach Expert Pedagogical Quality in Math Tutoring but Differ in Instructional and Linguistic Profiles
 }

\author{
\textbf{Ramatu Oiza Abdulsalam\textsuperscript{1}},  
\textbf{Segun Aroyehun\textsuperscript{2}},\\
\textsuperscript{1}African University of science and Technology \\
  Federal Capital Territory, Nigeria \\
  \texttt{rabdulsalam@student.aust.edu.ng}\\
  \textsuperscript{2}University of Konstanz \\
  Konstanz, Germany \\
  \texttt{segun.aroyehun@uni-konstanz.de} \\
    }

\begin{document}
\maketitle
\begin{abstract}
Recent work has explored the use of large language models (LLMs) to generate tutoring responses in mathematics, yet it remains unclear how closely their instructional behavior aligns with expert human practice. We analyze a dataset of math remediation dialogues in which expert tutors, novice tutors, and seven LLMs of varying sizes, comprising both open-weight and commercial models, respond to the same student errors.
We examine instructional strategies and linguistic characteristics of tutoring responses, including uptake (restating and revoicing), pressing for accuracy and reasoning, lexical diversity, readability, politeness, and agency. We find that expert tutors produce higher-quality responses than novices, and that larger LLMs generally receive higher pedagogical quality ratings than smaller models, approaching expert performance on average. However, LLMs exhibit systematic differences in their instructional profiles: they underuse discursive strategies characteristic of expert tutors while generating longer, more lexically diverse, and more polite responses.
Regression analyses show that pressing for accuracy and reasoning, restating and revoicing, and lexical diversity, are positively associated with perceived pedagogical quality, whereas higher levels of agentic and polite language are negatively associated. These findings highlight the importance of analyzing instructional strategies and linguistic characteristics when evaluating tutoring responses across human tutors and intelligent tutoring systems.

\end{abstract}
\section{Introduction}
Effective feedback plays a central role in supporting student learning by acknowledging effort, identifying errors, and providing clear guidance for self-correction \cite{lepper2002wisdom}. In instructional settings, the quality of feedback depends not only on whether mistakes are addressed, but also on how explanations are framed linguistically and pedagogically. As large language models (LLMs) are increasingly explored to generate tutoring responses \cite{wang2024bridging, pal2024autotutor}, an important open question is how closely their instructional behavior aligns with that of human tutors when responding to student errors.

Prior work \cite{zanotto-aroyehun-2025-linguistic, shaib-etal-2024-detection, namuduri2025qudsim} shows that LLM-generated text exhibits systematic linguistic regularities relative to human writing, including characteristic patterns of lexical choice and reduced stylistic variability. However, this line of work has largely examined language without reference to task-specific evaluation criteria, leaving it unclear whether such regularities influence instructional behavior or pedagogical evaluation in tutoring contexts. At the same time, educational research has examined effective feedback and instructional discourse, while computational linguistics has developed tools to measure linguistic properties such as readability, politeness, lexical diversity, and agency. These strands of work have largely evolved independently, and their connection in the context of tutoring remains underexplored.

We address this gap by analyzing an existing dataset \cite{kochmar-etal-2025-findings, maurya-etal-2025-unifying} in which expert tutors, novice tutors, and seven large language models respond to the same mathematics remediation prompts. We examine both instructional strategies and linguistic characteristics of tutor responses, including restating and revoicing, pressing for accuracy and reasoning, lexical diversity, readability, politeness, and agency. We evaluate responses in terms of pedagogical quality, based on structured annotations of error identification and guidance, and analyze how this quality relates to both instructional strategies and linguistic characteristics across human and LLM tutors.

Understanding these relationships is increasingly important as LLMs are explored for instructional support \cite{wang-etal-2025-training, team2025ai}. This raises questions about whether their responses exhibit linguistic characteristics associated with effective feedback and how such insights can help make LLMs more pedagogically aware and steerable.

This research is guided by three research questions.
\textbf{RQ1:} How do instructional strategies and linguistic characteristics differ across expert human tutors, novice human tutors, and LLMs in responses to the same mathematics remediation prompts?
\textbf{RQ2:} How does perceived pedagogical quality vary across expert human tutors, novice human tutors, and LLMs?
\textbf{RQ3:} Which instructional strategies and linguistic characteristics are associated with perceived pedagogical quality in tutoring responses.

\textbf{This paper makes three contributions.} First, we characterize systematic differences in instructional strategies and linguistic features across expert tutors, novice tutors, and LLMs responding to identical mathematics remediation prompts. Second, we compare perceived pedagogical quality across these tutor groups. Third, we identify which instructional strategies and linguistic features are associated with variation in pedagogical quality.

\section{Related Work}
Recent research has examined the use of LLMs as mathematical tutors, with mixed evidence regarding their pedagogical quality. While LLMs can generate fluent and well-structured feedback, human tutors continue to outperform them on core functions such as accurately identifying and correcting student errors, particularly in cases involving conceptual misunderstandings or multi-step reasoning \cite{wang2024bridging}. Related work also shows that LLMs may produce seemingly appropriate feedback while failing to correctly determine whether a response is incorrect \cite{kakarla2024using}.

Comparative studies further highlight differences in feedback structure: both human and LLM tutors rely on hint-based guidance, but LLMs more often provide compound feedback, whereas human tutors tend to deliver focused, single-action interventions \cite{kucheria2025comparing}. However, this work does not examine whether such differences correspond to systematic linguistic properties or how they relate to pedagogical quality.

A parallel line of research identifies stable linguistic differences between texts written by humans and those generated by LLMs, including lexical, syntactic, and discourse-level variation \cite{zanotto-aroyehun-2025-linguistic, shaib-etal-2024-detection, namuduri2025qudsim}. Yet this literature focuses on general writing tasks and does not address instructional feedback or pedagogical quality. Similarly, evaluations of tutoring interactions emphasize engagement, empathy, and conciseness \cite{chowdhury2025educators}, but typically assess these aspects independently of correctness and error identification.

Educational research suggests that effective feedback depends on accurately identifying and addressing student errors \cite{vanlehn2011relative, nicol2006formative, bamberger2010math}. Although LLMs can generate more extensive and readable feedback than human tutors \cite{rashid2024humanizing}, their tendency to produce fluent but incorrect responses raises concerns, particularly when learners treat fluency as a cue for correctness \cite{hattie2007power}. While LLMs can perform error detection and correction in isolation \cite{li2024evaluating}, they remain less effective than human tutors in adapting feedback to student misconceptions \cite{liu2023novice}. Related work on pragmatics and pedagogy \cite{pearson1995pragmatics} further highlights how conversational and politeness strategies may shape the clarity and interactional costs of tutoring, influencing both learner engagement and the effectiveness of corrective feedback.

Taken together, prior work has examined LLM tutoring performance, feedback structure, and linguistic differences, but these strands remain largely disconnected. In particular, there is limited empirical work linking measurable linguistic characteristics of tutoring responses to pedagogical quality in mistake remediation across both human and LLM tutors.

\section{Methodology}

\subsection{Data}
We use a dataset introduced as part of the BEA 2025 Shared Task on Pedagogical Ability Assessment of AI-powered Tutors \cite{maurya-etal-2025-unifying, kochmar-etal-2025-findings}. The dataset contains tutoring conversations based on student mathematical errors, where both AI systems and human tutors provide responses aimed at diagnosing and correcting the mistakes.
Our analysis uses 296 teacher-student dialogues at the middle-school
level \cite{macina2023mathdial} and the elementary level
\cite{wang2024bridging}. Each dialogue is paired with responses from
multiple tutors (humans and LLMs), resulting in 2{,}444 tutor
responses in total. Responses from the novice tutor are available for
76 dialogues. We report number of dialogues per tutor in
Table~\ref{tab:total_feedback} in the Appendix.

Each response includes annotations that identify pedagogically relevant features such as mistake identification, mistake location, providing guidance and actionability of the tutor's response \cite{kochmar-etal-2025-findings}. Each conversation features eight to nine responses generated from two  human tutors (an Expert and a Novice) and the remainder generated by large language models, including GPT-4 \cite{achiam2023gpt}, Gemini \cite{team2024gemini}, Sonnet (Anthropic), Mistral 7B \cite{jiang2023mistral}, Llama-3.1-405B  and two light weight models Llama-3.1-8B  \cite{grattafiori2024llama} and Phi-3 \cite{abdin2024phi}.
We provide an example prompt used to generate responses from LLM tutors in the Appendix (Figure \ref{fig:prompt-template}).
The structure of the dataset makes it well-suited for studying how human and LLM tutors differ in both linguistic expression (e.g., lexical complexity and  politeness) and pedagogical effectiveness, thus enabling systematic comparison.

\paragraph{Dataset Annotations}
The dataset consists of human annotation of responses using a set of pedagogical dimensions described in prior work \cite{kochmar-etal-2025-findings, maurya-etal-2025-unifying}. For completeness, we provide details for the annotated dimensions as follows:

\begin{itemize}
\item \textbf{Mistake Identification:} The tutor's response should identify the student's mistake or confusion.

\item \textbf{Mistake Location:} Tutor's response should clearly specify where the student's mistake is located in their response.

\item \textbf{Providing Guidance:} The tutor's response should contain explanatory guidance.

\item \textbf{Actionability:} The tutor's response should inform the student on what they should do next. 
\end{itemize}
The annotations were assigned using a three-level scale: Yes, To some extent, and No, indicating whether the tutor's response fully demonstrated, partially demonstrated, or did not demonstrate the pedagogical quality, respectively. For our analyses, we map the labels to numerical values (Yes = 2, To some extent = 1, No = 0). We then computed an overall pedagogical quality score by summing the values across the four evaluation dimensions.
\subsection{Linguistic and instructional features extraction}
We extract a set of instructional and linguistic features designed to capture complementary aspects of tutor responses to student mistakes. Together, these features characterize how explanations are constructed, both in terms of instructional moves and surface linguistic form.

\paragraph{Instructional Features}
Three instructional features capture tutors' use of pedagogically salient discourse moves derived from accountable talk theory \cite{michaels2010accountable}, which conceptualizes high-quality learning interactions as discourse that is accountable to knowledge, reasoning, and shared understanding. Given that the dataset analyzed here consists of dialogue-based interactions (including simulations with LLM tutors), we operationalize these dimensions at the level of paired exchanges between the preceding student turn and a tutor response, allowing us to capture not only correctness and reasoning but also local forms of uptake within the interaction \cite{suresh-etal-2022-talkmoves}. In this framework, effective tutors not only evaluate correctness but also elicit, probe, and build on learners' reasoning while remaining responsive to their contributions.

Pressing for accuracy indicates whether a response explicitly challenges or prompts the learner to reconsider the correctness of their answer, for example by questioning assumptions or highlighting inconsistencies. Pressing for reasoning captures whether the tutor prompts the learner to explain, justify, or elaborate on their thinking, thereby making underlying reasoning processes explicit. Uptake (as restating or revoicing) reflects whether the tutor reformulates the learner's response in their own words, a strategy that both clarifies understanding and demonstrates responsiveness to the learner's contribution.

These features are operationalized as probabilistic scores reflecting the likelihood of the respective instructional moves in each response, using a transformer-based text-pair classification model that takes the student turn and the corresponding tutor response as input.

We train a RoBERTa large model \cite{liu2019roberta} on the TalkMoves dataset \cite{suresh-etal-2022-talkmoves} to identify tutor discursive moves (see Table \ref{tab:model_test_result} for the performance metric of the model).

We use the probabilistic outputs of classifiers as continuous features to represent the degree to which each response exhibits a given instructional or linguistic feature, rather than converting classifier outputs to discrete labels. This allows us to capture graded variation in both binary and multiclass settings and avoids information loss from thresholding. This approach is consistent with recent work that leverages classifier probabilities as continuous indicators in downstream statistical analyses \cite{lasser2025collective}. Validation of this approach, including contrasting examples and additional human annotations, is provided in the Appendix (see Tables \ref{tab:classifier_eval} and \ref{tab:ex_response}). 

\paragraph{Linguistic Features}

Response length is measured as the total number of tokens in each response and log-transformed to reduce skew and limit the influence of outliers. Lexical diversity is quantified using the Measure of Textual Lexical Diversity (MTLD), which captures the range of vocabulary use independently of text length. MTLD provides an index of lexical variation that is well suited to short-to-medium instructional texts \cite{mccarthy2010mtld}.

Readability is assessed using the Flesch–Kincaid Reading Ease score \cite{kincaid1975derivation}, a widely used measure of textual accessibility based on the average number of words per sentence and average number of syllables per word. In the context of instructional feedback, readability serves as an indicator of surface-level processing demands imposed on the learner.

Readability and lexical diversity are used as proxies for instructional complexity at the linguistic level, capturing potential variation in cognitive processing demands from tutor responses.

Politeness is estimated using a transformer-based classifier \cite{srinivasan-choi-2022-tydip} trained to detect pragmatic markers of interpersonal tone, such as mitigation, respectfulness, and indirectness. This feature captures aspects of how corrective feedback is framed socially. We use the probabilistic output of the classifier corresponding to the politeness category as a feature.

Agency is operationalized using a transformer-based model that estimates the degree of agentic expression in text \cite{nikadon2025bertagent}, including linguistic cues related to intention, control, and action orientation. In tutoring responses, agency reflects how explanations position the learner and the tutor with respect to problem-solving and corrective action. We use the probabilistic output of the classifier as a feature. 

All features are computed at the level of individual tutor responses.

\subsection{Analytic strategy}

Our analytic strategy was designed to address the three main research questions concerning (i) linguistic differences between human and LLM tutors, (ii) differences in pedagogical quality across sources, and (iii) the relationship between linguistic traits and pedagogical quality.

First, to examine linguistic and instructional differences between human and LLM tutors (RQ1), we compared tutor responses across eight features: pressing for reasoning, pressing for accuracy, restating/revoicing, response length (log transformed), lexical diversity (MTLD), readability (Flesch–Kincaid Reading Ease), politeness, and agency. For each feature, we estimated group-level mean differences relative to expert human tutors, which serve as a baseline. We provide example tutor responses where the features are salient in Appendix \ref{sec:appendix_examples}.

Second, to assess pedagogical quality across tutor types (RQ2), we relied on the four-dimensional annotation framework capturing mistake identification, mistake location, actionability, and provision of guidance. We aggregated scores assigned to each pedagogical dimension to derive a composite pedagogical quality score.

Third, to examine how linguistic traits are associated with pedagogical quality (RQ3), 
we construct a turn-centered measure of pedagogical quality. For each conversation turn, we compute the average pedagogical quality score across all tutors who responded to that turn and express each tutor’s score as a deviation from this turn-level mean. This transformation yields a measure of relative pedagogical quality that captures how a tutor’s response compares to alternative responses to the same instructional prompt.

We then summarize relative pedagogical quality at the tutor level by computing the mean deviation across all turns answered by a given tutor. Uncertainty estimates are obtained by computing standard errors across conversation turns. By anchoring comparisons within turns, this approach controls for variation in turn difficulty and isolates differences attributable to tutors' responses rather than to the instructional context itself.
While the turn-centered measure provides a descriptive comparison of tutors' relative pedagogical quality, it does not address how specific linguistic characteristics are associated with variation in pedagogical quality within turns.

To examine these associations, we estimated a linear fixed effects model at the conversation level. By demeaning both the outcome and predictors within each conversation, this specification isolates within-conversation variation, thereby controlling for all unobserved characteristics of the instructional context, such as task difficulty or error type. 

All predictors were standardized for comparability, and standard errors were clustered at the conversation level to account for dependence among responses within the same interaction.
This combined approach allows us to assess raw patterns in the data while also estimating adjusted relationships between linguistic features and pedagogical quality.

\begin{figure*}[!ht]
    \centering  \includegraphics[width=\textwidth]{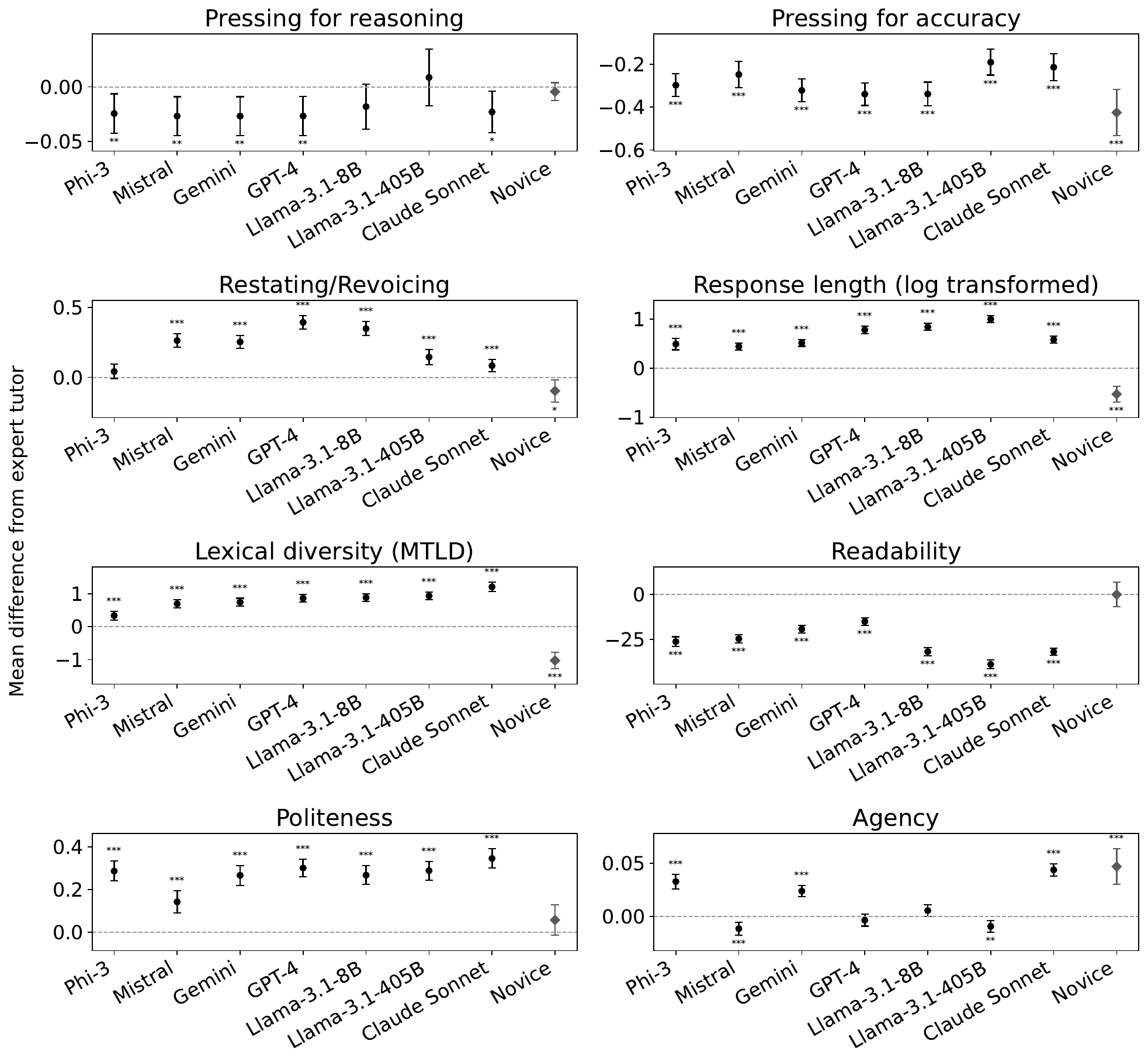}
    \caption{Instructional and linguistic profiles of tutors relative to expert tutor baseline.
Each panel shows the mean difference between a tutor and the expert tutor for a given feature, with error bars indicating 95\% confidence intervals. Positive values indicate higher feature values than the expert tutor, while negative values indicate lower values. The horizontal dashed line denotes parity with the expert tutor. Note: * indicates p < .05, ** indicates p < .01, and *** indicates p < .001 after Benjamini–Hochberg false discovery rate correction.
}
    \label{fig:ling}
\end{figure*}
\begin{figure*}[!htbp]
    \centering  \includegraphics[width=\textwidth]{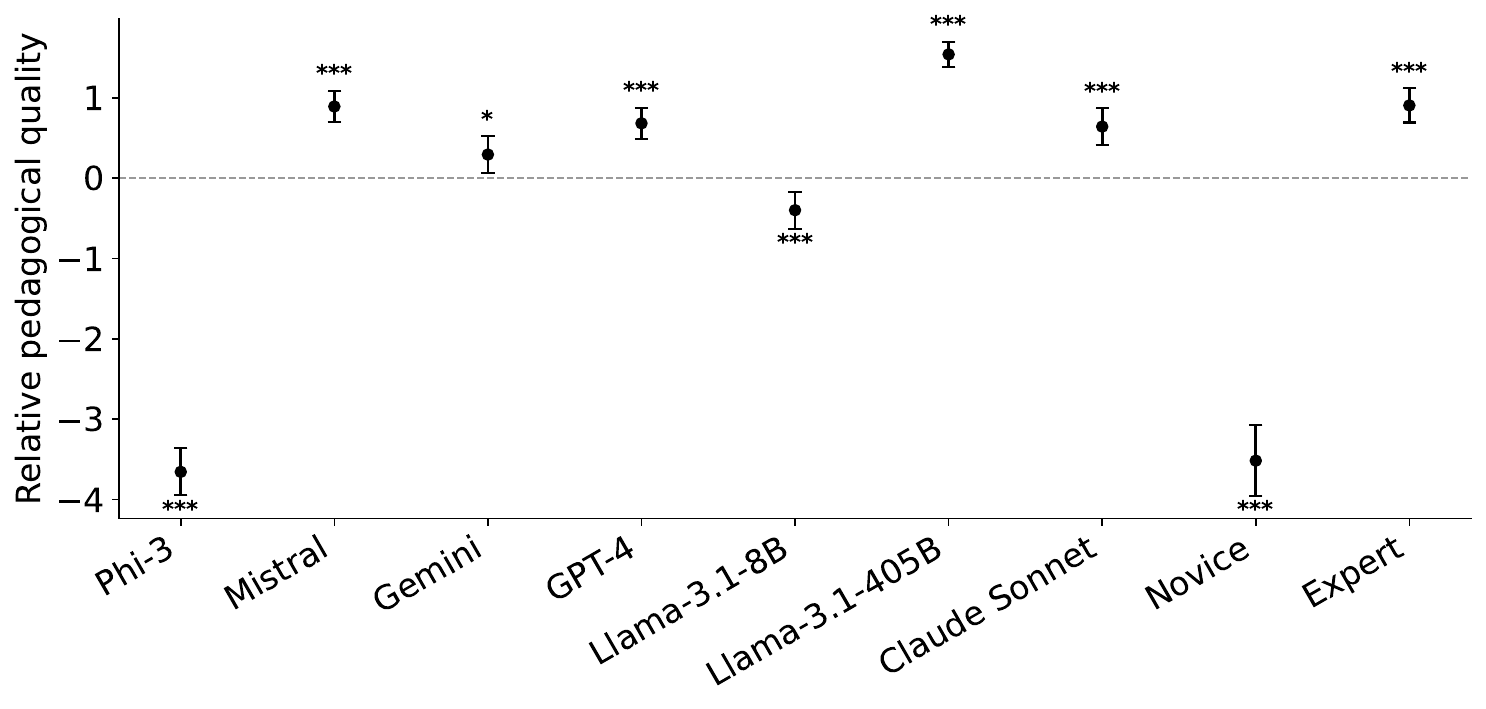 }
    \caption{Relative pedagogical quality across tutors. Each datapoint shows average relative pedagogical quality for each tutor (with 95\% CI) when responding to the same conversation turns. The dashed line indicates parity with the turn-level average; values above (below) zero indicate higher (lower) perceived pedagogical quality relative to other tutors. Note: * indicates p < .05, ** indicates p < .01, and *** indicates p < .001 after Benjamini–Hochberg false discovery rate correction.}
    \label{fig:relativequality}
\end{figure*}

\section{Results}

\subsection{RQ1: Linguistic Differences Between Human and LLM Tutors}
Figure \ref{fig:ling} compares human and LLM tutors along eight instructional and linguistic dimensions, expressed as mean differences relative to expert human tutors, with error bars indicating uncertainty around the estimates. By construction, positive values indicate higher scores than expert tutors, whereas negative values indicate lower scores.

Across most dimensions, LLM tutors exhibit systematic and internally consistent differences relative to expert human tutors. For instructional moves, LLMs generally show lower levels of pressing for reasoning and pressing for accuracy than expert tutors.
In contrast, restating or revoicing is consistently more prevalent in LLM-generated responses, with all models exhibiting negative deviations relative to expert tutors. Novice human tutors show lower levels of restating/revoicing, suggesting that this instructional move may be less characteristic of expert tutoring and general feature of human responses.

LLM responses are substantially longer than those produced by expert tutors, as reflected in positive deviations on log-transformed response length. This pattern is consistent across models, whereas novice human tutors produce markedly shorter responses than experts. A similar ordering is observed for lexical diversity, where LLMs exhibit higher MTLD scores than expert tutors, while novice tutors show substantially lower lexical diversity.

Differences in readability are pronounced. All LLM tutors score substantially lower on the Flesch–Kincaid Reading Ease metric relative to expert tutors, indicating that their responses are, on average, less readable. The magnitude of this difference varies across models but is consistently negative. Novice tutors, by contrast, are closer to experts on readability, with estimates centered near zero.

LLM tutors also exhibit higher politeness scores than expert tutors across models, whereas novice tutors show only modest deviations. This suggests that elevated politeness is a stable feature of LLM-generated feedback rather than a general property of non-expert tutoring.

Finally, patterns for agency differ from other linguistic features. Some LLMs exhibit lower agentic expression than expert tutors, while others are closer to parity. Novice tutors show higher agency scores than experts, suggesting that agentic language does not align with human tutoring expertise.

Taken together, these results indicate that LLM tutors differ from human tutors along multiple linguistic and instructional dimensions, with consistent differences in pressing for reasoning, pressing for accuracy, restating/revoicing, response length, lexical diversity, readability, and politeness.

\subsection{RQ2: Pedagogical Quality Across Tutor Types}
Figure \ref{fig:relativequality} reports tutors' average relative pedagogical quality scores, computed as deviations from the turn-level mean across all responses to the same student turn. By design, positive values indicate responses that score higher than other responses to the same instructional prompt, whereas negative values indicate lower relative pedagogical quality.

Clear differences emerge across tutor types. Expert human tutors exhibit consistently positive relative pedagogical quality, indicating that their responses tend to score higher than other responses to the same student mistakes. In contrast, novice human tutors display clear negative deviations, suggesting substantially lower pedagogical quality when evaluated relative to alternative responses within the same conversational context.

LLM tutors span a wide range of relative pedagogical quality. Some models cluster near or above zero, indicating performance comparable to or exceeding the turn-level average, whereas others fall below zero. Notably, higher-performing LLMs approach the relative pedagogical quality of expert tutors on average, while lower-performing model (Phi-3) exhibit negative deviation similar to novice tutors.
Importantly, these differences are observed within identical instructional contexts and therefore reflect variation in tutor responses rather than differences in student turns or task difficulty. We report in the Appendix, Figure \ref{fig:relative_dimension}, the patterns observed within each dimension with some heterogeneity across pedagogical dimensions. While Llama-3.1-405B, Mistral, and Claude Sonnet perform consistently well across dimensions, other models exhibit more uneven performance. For example, GPT-4 performs strongly on diagnostic dimensions and guidance but comparatively weaker on actionability, whereas the expert tutor shows its strongest relative advantage on actionability. 

\subsection{RQ3: Associations Between Linguistic and Instructional Features and Pedagogical Quality}

\begin{figure*}[!ht]
    \centering  \includegraphics[width=\textwidth]{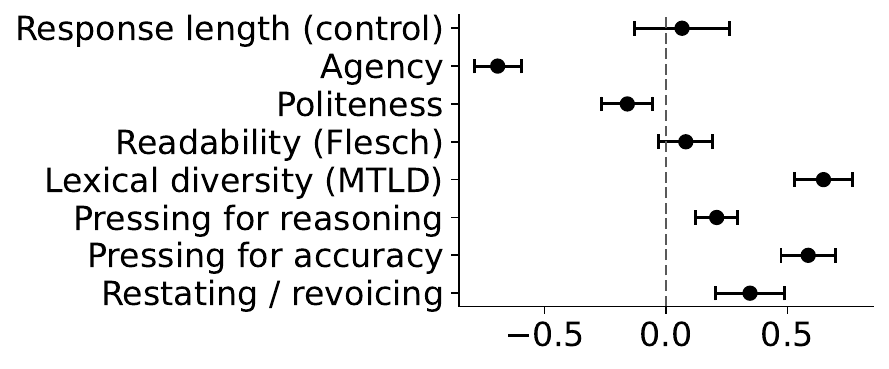 }
    \caption{Instructional and linguistic correlates of pedagogical quality. Coefficients from an ordinary least squares model predicting perceived pedagogical quality at the tutor response level. Horizontal lines indicate 95\% confidence intervals. The specification includes a control for response length. Positive coefficients indicate that higher values of a feature are associated with higher perceived pedagogical quality, while negative coefficients indicate associations with lower perceived pedagogical quality. Standard errors are clustered by conversation turn to account for multiple tutor responses to the same turn.
}
    \label{fig:olsquality}
\end{figure*}

Figure \ref{fig:olsquality} reports standardized coefficients from a within-conversation regression predicting relative pedagogical quality from our set of instructional and linguistic features. Additional details on the regression model results are reported in Table \ref{tab:rq3_regression} in the Appendix.

All three instructional moves are positively associated with pedagogical quality.
Pressing for accuracy shows a large positive association ($\beta = 0.586$, 95\% CI [0.474, 0.698], p < 0.001), followed by uptake (restating/revoicing) ($\beta = 0.346$, 95\% CI [0.205, 0.488], p < 0.001). Pressing for reasoning also shows a positive association ($\beta = 0.209$, 95\% CI [0.122, 0.295], p < 0.001).

Among linguistic variables, lexical diversity (MTLD) is positively associated with pedagogical quality ($\beta = 0.649$, 95\% CI [0.528, 0.770], p < 0.001). In contrast, readability (Flesh reading ease) is not statistically associated with pedagogical quality ($\beta = 0.082$, 95\% CI [-0.029, 0.192], p = 0.149), and response length is not statistically associated with pedagogical quality ($\beta = 0.066$, 95\% CI [-0.128, 0.261], p = 0.505).

Two linguistic variables are negatively associated with pedagogical quality. Politeness shows a negative association ($\beta = -0.160$, 95\% CI [-0.265, -0.055], p = 0.003), as does agency ($\beta = -0.693$, 95\% CI [-0.790, -0.596], p < 0.001).

In sum, pedagogical quality is positively associated with
pressing for reasoning, pressing for accuracy,restating/revoicing,  and lexical diversity, while politeness and agency show negative association. Readability and response length show no statistically detectable association in this specification.
To assess whether these associations are driven by a single dimension or are consistent across pedagogical subdimensions, Figure \ref{fig:coef_dimensions} (in the Appendix) shows estimates of the same model specification separately for each dimension. The dimension-level models broadly support the aggregate pattern: lexical diversity and most pedagogical moves are consistently positively associated with pedagogical quality. One exception is uptake (restating/revoicing), which is not significantly associated with the actionability dimension. In contrast, agency remains consistently negatively associated across dimensions, while politeness retains a negative association only for actionability.

\section{Discussion}
The positive associations between pedagogical quality and features such as restating and revoicing, pressing for accuracy and reasoning, and lexical diversity point to the role of cognitively engaging instructional strategies. Restating and revoicing can support alignment and comprehension by making student reasoning explicit, while pressing for accuracy and reasoning encourages learners to diagnose and repair their own errors. These patterns are consistent with prior work emphasizing active engagement and self-explanation in effective learning \cite{chi2001learning, graesser1995collaborative, vanlehn2011relative}. Lexical diversity may similarly reflect more elaborated explanations that provide richer cues for understanding and correction.

In contrast, the negative associations for agentic language and politeness suggest a distinction between supportive stance-taking and pragmatically actionable instructional guidance. Prior work on pragmatics and pedagogy~\cite{pearson1995pragmatics} suggests that conversational politeness strategies and socially accommodating dialogue patterns can sometimes inhibit effective tutoring by weakening corrective feedback or obscuring instructional signals. This distinction is visible in a tutor response such as "That is a good try" which acknowledges learner effort but provides little information for diagnosing or repairing student error. In contrast, a tutor response such as "Great job working through the problem step-by-step, but let's take a closer look at that final calculation..." combine supportive framing with explicit corrective guidance that directs the learner's attention towards the relevant misconception and remediation process. Recent work on large language models similarly highlights tendencies toward overly accommodating or sycophantic responses \cite{fanous2025syceval, cheng2026elephant}, which may reduce the clarity and directness of corrective feedback in instructional settings.

\section{Conclusion}
This study provides an empirical analysis of pedagogical quality in human and LLM-based tutoring by jointly examining instructional moves and linguistic features while holding conversational context constant. Our findings show that pedagogical quality is most strongly associated with specific instructional moves i particular pressing for reasoning, pressing for accuracy, and restating/revoicing learner responses—alongside a limited set of linguistic characteristics.

Among linguistic features, lexical diversity is positively associated with pedagogical quality, whereas politeness and agentic language are negatively associated; readability and response length show no detectable effects. These results suggest that effective feedback depends less on linguistic complexity or assertiveness than on how linguistic choices align with instructional function.

Both human and LLM tutors exhibit substantial variation along these dimensions, with higher-quality responses combining targeted instructional moves and supportive linguistic configurations, regardless of tutor type. This indicates that pedagogical quality is best understood as a property of response-level features rather than tutor identity.

For LLM-based tutors, these findings suggest that optimization should focus on reproducing feature combinations associated with higher pedagogical quality, rather than increasing linguistic expressiveness or mimicking human tutors.
More broadly, our approach may extend to other domains of human–AI interaction.


\section{Limitations}
Given the scope of this study, several limitations should be acknowledged. First, although the dataset comprises 2,444 tutor responses (and a maximum of 296 conversations per tutor), which is sufficient for the within-turn comparative analyses conducted here, a larger dataset would allow for more precise estimation of smaller effects and finer-grained analyses across tutor groups. Second, our analyses rely on a dataset in which human and model responses are generated from the same underlying student inputs, enabling direct comparison under matched conditions. However, LLM responses depend on the specific prompting setup used to elicit them, and alternative prompt formulations may produce different outputs. The findings should therefore be interpreted as reflecting model behavior under the prompt configuration examined here rather than the upper bound of LLM capabilities

Third, the set of instructional and linguistic features examined in this study is necessarily selective rather than exhaustive. The features were chosen to capture key instructional moves and relevant dimensions of linguistic expression, other aspects of tutor responses such as discourse structure, epistemic framing, or mathematical specificity may also be relevant to pedagogical quality and warrant systematic investigation in future work.

Fourth, the analysis is limited to English language interactions in a mathematical tutoring context. Although some of the instructional patterns identified here may generalize to other domains or languages, extending this framework to additional subject areas and linguistic contexts is an important direction for future research.

Finally, pedagogical quality in this study is assessed through structured annotations (and does not aim to be exhaustive) applied to single-turn tutor responses. While this approach allows for controlled comparison across identical instructional prompts, it does not capture downstream student learning outcomes. Future work should extend pedagogical quality to multi-turn interactions and incorporate student responses or learning signals as complementary indicators of pedagogical effectiveness.
%
%
\bibliography{custom}

\appendix{}
\section{Dataset}
Table \ref{tab:total_feedback} shows the number of responses for each tutor in the dataset.

\begin{table}[!ht]
 \begin{tabular}{lc}
\hline
\textbf{Tutor} & \textbf{Number of responses} \\
\hline

Expert & 296 \\
Novice & 76 \\
GPT-4 & 296 \\
Gemini & 296 \\
Sonnet & 296 \\
Mistral 7B & 296 \\
Llama-3.1-405B & 296 \\
Llama-3.1-8B  & 296 \\
Phi-3 & 296 \\ \hline
Total & 2444 \\

\hline
\end{tabular}
\caption{Number of responses per tutor}
\label{tab:total_feedback}
\end{table}

\section{Classifier evaluation}
For each feature, we select 25 responses with the highest predicted probabilities and the 25 responses with the lowest predicted probabilities (for a total of 50). These responses are manually annotated and treated as the reference labels. We then evaluate the correspondence between classifier scores and human annotations using the area under the receiver operating characteristic curve (AUC), which assesses the extent to which higher classifier probabilities are assigned to human-labeled positive cases relative to negative cases. We use AUC because it evaluates ranking quality independently of a fixed classification threshold and is therefore well suited for continuous probabilistic scores. Similar validation approach has been used in prior work examining automatically derived continuous measures \cite{gennaro2022emotion, Aroyehun2024Computational}. Table \ref{tab:classifier_eval} shows the results.

\begin{table}[!ht]
 
 \begin{tabular*}{\columnwidth}{l @{\extracolsep{\fill}} c}
\hline
\textbf{Feature} & \textbf{AUC Score} \\
\hline
Politeness & 0.738 \\
Agency & 0.740 \\
PressAccuracy & 0.614 \\
Uptake & 0.827 \\
PressReasoning & 0.794 \\

\hline
\end{tabular*}
\caption{Classifiers were evaluated by comparing the AUC scores of the top  and bottom  classifier-generated probabilities, using human annotations as the ground truth.}

\label{tab:classifier_eval}
\end{table}

\section{Performance of fine-tuned transformers model}
We train a RoBERTa large model on the TalkMoves dataset to identify tutor discursive move. Table \ref{tab:model_test_result} reports the performance of the model on the unseen test set. Note that we only consider the relevant discursive moves for this study, namely: pressing for accuracy, pressing for reasoning, and uptake (restating/revoicing).

\begin{table*}[!htbp]
\centering
 \begin{tabular}{lccccc}
\hline
\textbf{Label} & \textbf{Precision} & \textbf{Recall} & \textbf{F1} & \textbf{Support} \\
\hline

Nomove & 0.963 & 0.895 & 0.928 & 9123  \\

Pressing for accuracy & 0.777 & 0.903 & 0.836 & 2014  \\

Pressing for reasoning & 0.693 & 0.821 & 0.752 & 179  \\

Uptake (restating/revoicing) & 0.603 & 0.761  & 0.673 & 330  \\

Participation management & 0.694 & 0.778 & 0.733 & 1614  \\

\hline
\end{tabular}

\caption{Performance on the TalkMove test set of a fine-tuned RoBERTa-large model}
\label{tab:model_test_result}
\end{table*}

\section{Example text with extracted features}
\label{sec:appendix_examples}
Examples of tutors with top linguistic Scores include:

\begin{itemize}
\item\textbf{Politeness- Human Tutor:}
    \textit{“Can I ask what you thought about the third year for?”}
    
\item\textbf{Agency- Human Tutor:}
    \textit{”Great try! What was your first step?”}
    
\item\textbf{MTLD- Human Tutor:}
    \textit{"There seems to be mistake in here. could you please tell me how much money would it cost for a person to go through arcade tokens, mini golf and go-karting?"}
    
\item\textbf{Reading Ease- Human Tutor:}
    \textit{"Ah, not quite. 100 * 2 = 200"}
    
\item\textbf{Grade Level- Human Tutor:}
    \textit{"So, the two figures you have calculated are 16mph and 16 weeks, assuming a gain of an equal amount of speed in each week, how would you calculate the increase in speed in each week?"}
    
\item\textbf{Pressing for Reasoning- Human Tutor:}
    \textit{"Why are you adding 10mph?"}
    
\item\textbf{Pressing for Accuracy- Human:}
    \textit{"Great. If we want to find the difference between the 2 types of baked goods what do I need to do first"}
    
\item\textbf{Uptake (Restating/Revoicing)- Human:}
    \textit{"Oh you're very close. 62 times 5 = 310, but 72 times 5 is a little bigger"}

\item\textbf{Politeness- LLM:}
    \textit{"Great job summarizing the main points of the lesson! Let's review any questions or areas you'd like to clarify further."}
    
\item\textbf{Agency- LLM:}
    \textit{"Great job on your progress; let's tackle the next challenge together!"}
    
\item\textbf{MTLD- LLM:}
    \textit{”Actually, let's take another look at that calculation together - I think there might be a small mistake in the multiplication, could you try it again or show me the exact steps your calculator is taking?"}
    
\item\textbf{Reading Ease- LLM:}
    \textit{"That's a good try, but let's try again. If we subtract 156 from 163, we get 7. Can you see how we got that?"}
    
\item\textbf{Reading Ease- LLM:}
    \textit{"Actually, I think there might be a small mistake in the last step of your calculation - 100 cents is indeed equal to \$1, but you might want to consider that the question asks for the cost in dollars, and your calculation is correct, so the final answer is correct, but let's make sure you understand that 100 cents equals 1 dollar, so in this case, the answer is indeed \$1, but in general, if you had more than 100 cents, you would need to convert, for example, 200 cents would be \$2, does that make sense?"}
    
\item\textbf{Pressing for Reasoning- LLM:}
    \textit{"Great job showing your work, and I see you're on the right track, but could you explain why you added 6 to 18 in your third step"}

\item\textbf{Pressing for Accuracy- LLM:}
    \textit{"That's not quite right; let's try adding them again together â€“ when we add 2700 and 1600, what do we get this time?"}
    
\item\textbf{Uptake (Restating/Revoicing)- LLM:}
    \textit{"Actually, the correct product of 7 multiplied by 60 is 420, not 520."}
\end{itemize}

\section{Contrasting examples between minimum and maximum scores based on classifiers}
Table \ref{tab:ex_response} presents examples of tutor responses with low and high classifier scores for agency, politeness, pressing for accuracy, pressing for reasoning, and uptake. The contrast between minimum and maximum scores provides qualitative face validity that the classifiers are capturing the intended constructs.

\begin{table*}[!ht]
\centering
\resizebox{\textwidth}{!}{
 \begin{tabular}{lp{7cm}p{7cm}}
\hline
\textbf{Feature} & \textbf{\makecell{Minimum score}} & \textbf{\makecell{Maximum score}} \\
\hline
Politeness    & \textbf{Response:} Talk me through why you subtracted 60 from 100? \newline \textbf{Score:} 0.0016 & \textbf{Response:} Great job summarizing the main points of the lesson! Let's review any questions or areas you'd like to clarify further. \newline \textbf{Score:} 0.9980 \\
\hline

Agency   & \textbf{Response:} Sorry, you have messed up \newline \textbf{Score:} 0.3921 & \textbf{Response:} Great job on your progress; let's tackle the next challenge together! \newline \textbf{Score:} 0.6750 \\
\hline

Pressing for reasoning    & \textbf{Response:} That was a good try! \newline \textbf{Score:} 0.0001 & \textbf{Response:} Why are you adding 10mph? \newline \textbf{Score:} 0.9932 \\
\hline

Pressing for accuracy   & \textbf{Response:} Your answer is incorrect. \newline \textbf{Score:} 0.0008 & \textbf{Response:} Great. If we want to find the difference between the 2 types of baked goods what do I need to do first? \newline \textbf{Score:} 0.9961 \\
\hline

Uptake (restating/revoicing)    & \textbf{Response:} Great job on completing that task! Can you now tell me what answer you got for 6 + 8? \newline \textbf{Score:} 0.0002 & \textbf{Response:} Actually, the correct product of 7 multiplied by 60 is 420, not 520. \newline \textbf{Score:} 0.9907 \\
\hline

\end{tabular}%
}
\caption{Examples of responses with minimum and maximum feature scores}
\label{tab:ex_response}
\end{table*}

\section{Regression model results}
\begin{figure*}
    \centering
    \includegraphics[width=\linewidth]{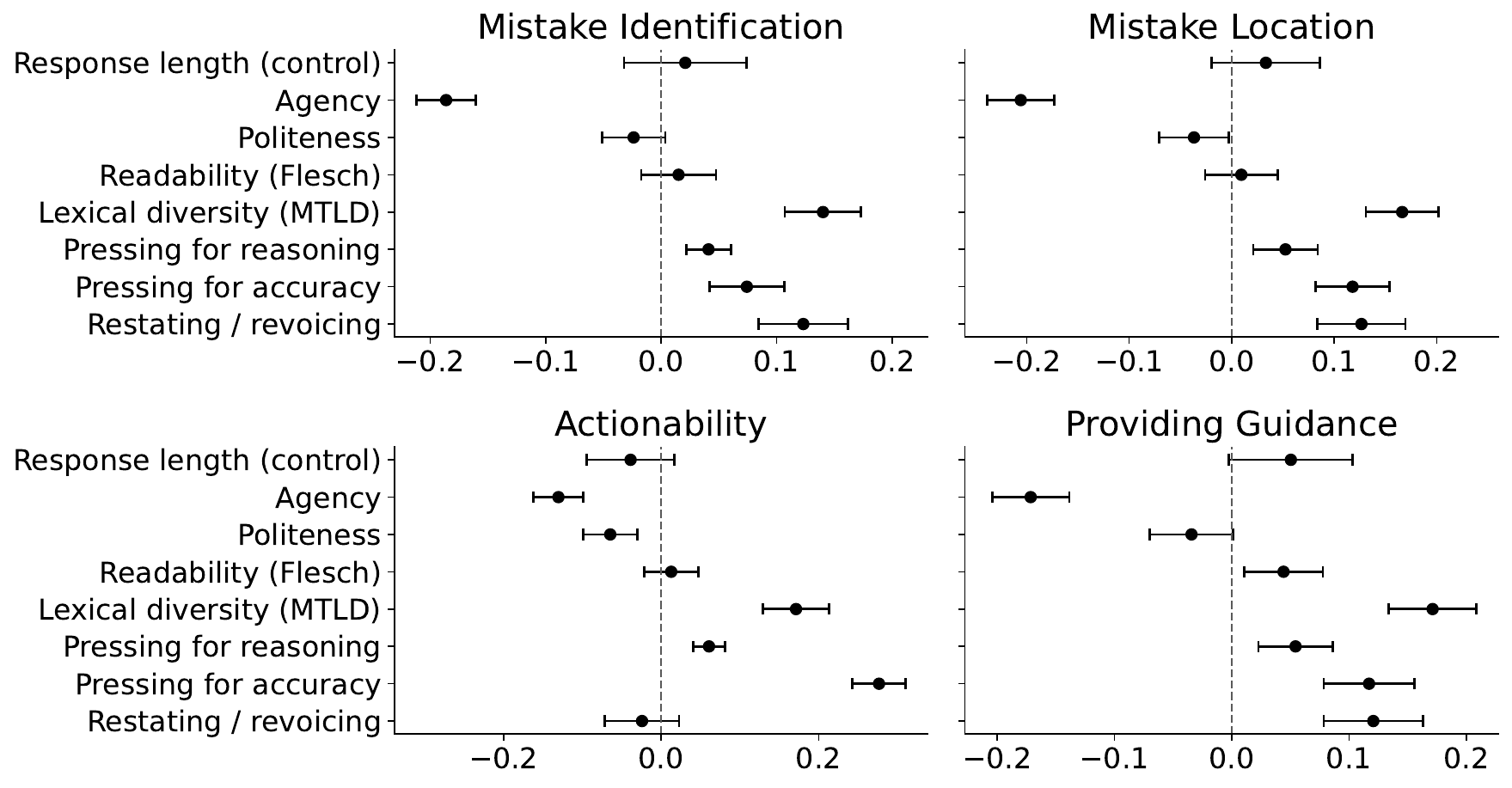}
    \caption{Regression coefficients for models predicting relative pedagogical quality for each dimension from instructional and linguistic features. Coefficients are standardized. Standard errors are clustered by conversation.}
    \label{fig:coef_dimensions}
\end{figure*}

Table \ref{tab:rq3_regression} shows the output of the regression model assessing the relationship between instructional moves and linguistic features and pedagogical quality. Figure \ref{fig:coef_dimensions} shows the corresponding relationship for each pedagogical dimension.
\begin{table*}[htbp]
\centering

\begin{tabular}{lccc}
\hline
Predictor & $\beta$ & 95\% CI & $p$ \\
\hline
Pressing for reasoning  & 0.209  & [0.122, 0.295]   & $< .001$ \\
Pressing for accuracy   & 0.586  & [0.474, 0.698]   & $< .001$ \\
Uptake (restating/revoicing)          & 0.346  & [0.205, 0.488]   & $< .001$ \\
Response length (log)      & 0.066  & [-0.128, 0.261]  & 0.505 \\
Lexical diversity (MTLD)             & 0.649  & [0.528, 0.770]   & $< .001$ \\
Flesch reading ease & 0.082  & [-0.029, 0.192]  & 0.149 \\
politeness     & -0.160 & [-0.265, -0.055] & 0.003 \\
Agency         & -0.693 & [-0.790, -0.596] & $< .001$ \\
\hline
Observations ($N$)   & \multicolumn{3}{r}{2444} \\
$R^2$                & \multicolumn{3}{r}{0.239} \\
Adjusted $R^2$       & \multicolumn{3}{r}{0.237} \\
F-statistic          & \multicolumn{3}{r}{110.5} \\
Prob (F-statistic)   & \multicolumn{3}{r}{$< .001$} \\
\hline
\end{tabular}

\caption{Within-conversation regression predicting relative pedagogical quality from instructional and linguistic features. Coefficients are standardized. Standard errors are clustered by conversation.}
\label{tab:rq3_regression}

\end{table*}

\section{Patterns within each pedagogical dimension}
\begin{figure*}
    \centering
    \includegraphics[width=\linewidth]{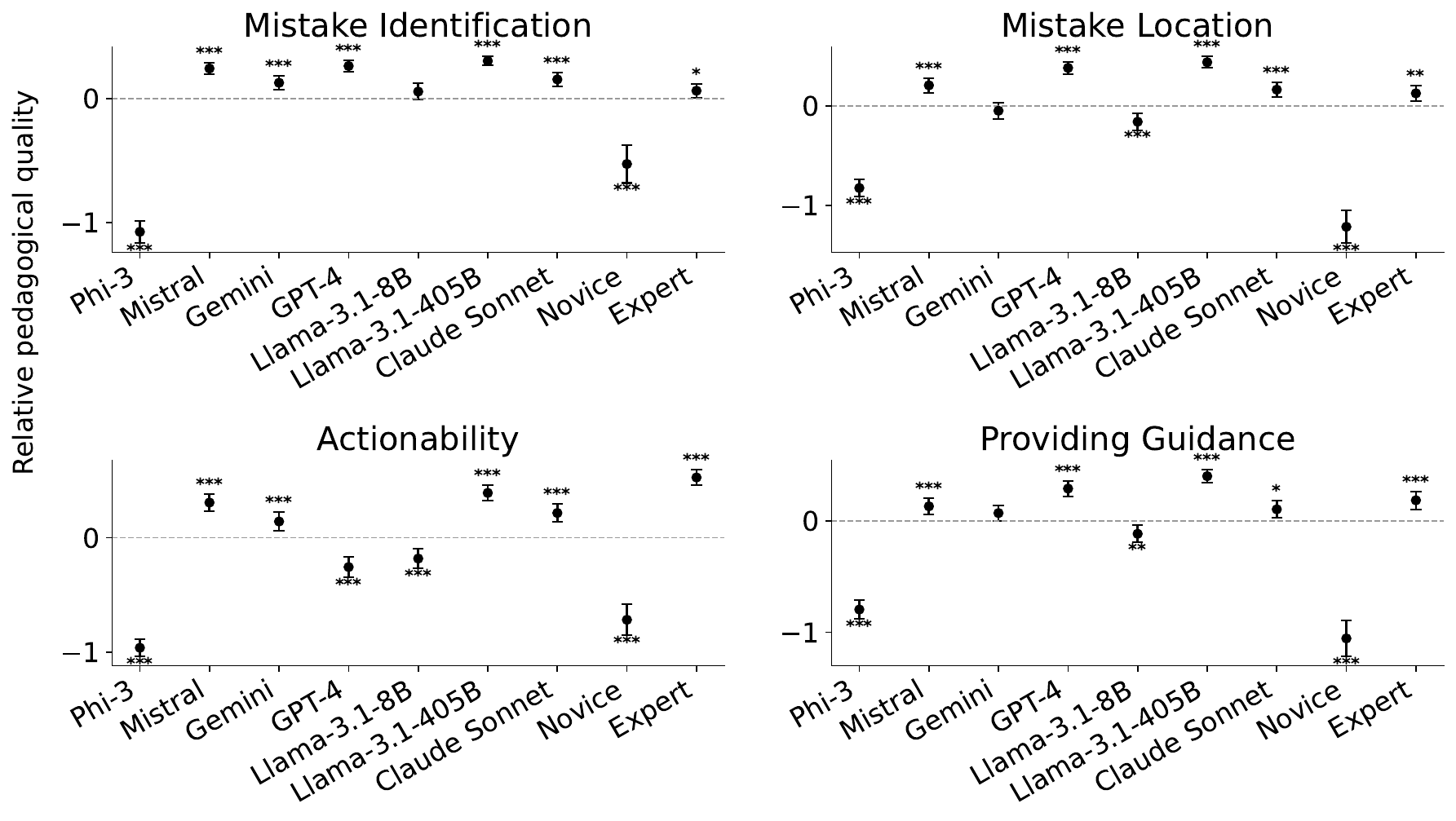}
    \caption{Relative pedagogical quality for each dimension across tutors. Each datapoint shows average relative pedagogical quality for
each tutor (with 95\% CI) when responding to the same conversation turns. The dashed line indicates parity with the
turn-level average; values above (below) zero indicate higher (lower) perceived pedagogical quality relative to other
tutors. Note: * indicates p < .05, ** indicates p < .01, and *** indicates p < .001 after Benjamini–Hochberg false discovery rate correction.}
    \label{fig:relative_dimension}
\end{figure*}

We report the dimension-level patterns in Figure \ref{fig:relative_dimension}. Performance differences are not uniform across dimensions. For example, GPT-4 performs strongly on diagnostic dimensions (mistake identification and mistake location) and guidance, but comparatively weaker on actionability. Llama-3.1-405B shows consistently positive performance across all four dimensions, with Mistral and Claude Sonnet displaying similar but more moderate patterns. The decomposition also reveals dimension-specific weaknesses obscured in the aggregate: Gemini is not significantly different on mistake location or guidance despite being slightly positive overall, while Llama-3.1-8B shows no significant difference on mistake identification but negative performance on the remaining dimensions. Finally, although the expert tutor performs above average overall, its advantage is strongest on actionability, whereas other models exhibit stronger relative performance on other pedagogical dimensions.

\section{Prompt Example}
Figure \ref{fig:prompt-template} provides an example prompt used to generate responses from LLM tutors.

\begin{figure*}[!htbp]
 \centering   

\begin{tcolorbox}[colback=gray!5,colframe=gray!40]
System: You are an experienced elementary school math teacher, and you are going to respond to a student's mistake in a useful and caring way 
  
 User: The problem your student is solving is on the topic: {Mensuration}
 
 Conversation History: {"Tutor: Finally, we need to add the areas of both rectangles.
 Tutor: Go ahead and try to find the answer.
 Tutor: What is the value of 36+42?
 Student: 1,512"} 
  
 Assistant: Tutor response (maximum one sentence that is most appropriate given topic and conversation history)
\end{tcolorbox}
\caption{Example prompt used to generate responses from large language models (source: \citet{kochmar-etal-2025-findings}). }
\label{fig:prompt-template} 

\end{figure*}

\section{Information on AI Usage}
Artificial intelligence tools were used to assist in writing source code and in editing and proofreading portions of this manuscript. All conceptual contributions and methodological decisions are those of the authors, who take full responsibility for the final content of the paper.
\end{document}